\documentclass[letterpaper, 10 pt, conference]{ieeeconf}  % Comment this line out if you need a4paper

\usepackage{graphicx}
\usepackage{multirow}

\IEEEoverridecommandlockouts                              % This command is only needed if 
\usepackage{amsmath} % assumes amsmath package installed
\usepackage{amssymb}  % assumes amsmath package installed

\newcommand{\eg}{\textit{e.g. }}
\newcommand{\etc}{\textit{etc.}}
\newcommand{\etal}{\textit{et al. }}
\newcommand{\ie}{\textit{i.e. }}

\title{\LARGE \bf
Towards End-to-End Lane Detection: an Instance Segmentation Approach
}
\author{Davy Neven \qquad Bert De Brabandere \qquad Stamatios Georgoulis \qquad Marc Proesmans \qquad Luc Van Gool\\
ESAT-PSI, KU Leuven\\
{\tt\small firstname.lastname@esat.kuleuven.be}
}

\begin{document}

\maketitle
\thispagestyle{empty}
\pagestyle{empty}

%%%%%%%%%%%%%%%%%%%%%%%%%%%%%%%%%%%%%%%%%%%%%%%%%%%%%%%%%%%%%%%%%%%%%%%%%%%%%%%%
\begin{abstract}

%This electronic document is a ÒliveÓ template. The various components of your paper [title, text, heads, etc.] are already defined on the style sheet, as illustrated by the portions given in this document.

Modern cars are incorporating an increasing number of driver assist features, among which automatic lane keeping.
The latter allows the car to properly position itself within the road lanes, which is also crucial for any subsequent lane departure or trajectory planning decision in fully autonomous cars.
Traditional lane detection methods rely on a combination of highly-specialized, hand-crafted features and heuristics, usually followed by post-processing techniques, that are computationally expensive and prone to scalability due to road scene variations.
More recent approaches leverage deep learning models, trained for pixel-wise lane segmentation, even when no markings are present in the image due to their big receptive field.
Despite their advantages, these methods are limited to detecting a pre-defined, fixed number of lanes, \eg ego-lanes, and can not cope with lane changes.
In this paper, we go beyond the aforementioned limitations and propose to cast the lane detection problem as an instance segmentation problem -- in which each lane forms its own instance -- that can be trained end-to-end.
To parametrize the segmented lane instances before fitting the lane, we further propose to apply a learned perspective transformation, conditioned on the image, in contrast to a fixed "bird's-eye view" transformation. 
By doing so, we ensure a lane fitting which is robust against road plane changes, unlike existing approaches that rely on a fixed, pre-defined transformation.
In summary, we propose a fast lane detection algorithm, running at 50 fps, which can handle a variable number of lanes and cope with lane changes. 
We verify our method on the tuSimple dataset and achieve competitive results.

\end{abstract}

%%%%%%%%%%%%%%%%%%%%%%%%%%%%%%%%%%%%%%%%%%%%%%%%%%%%%%%%%%%%%%%%%%%%%%%%%%%%%%%%
\section{INTRODUCTION}

Fully autonomous cars are the main focus of computer vision and robotics research nowadays, both at an academic and industrial level.
The goal in each case is to arrive at a full understanding of the environment around the car through the use of various sensors and control modules.
Camera-based lane detection is an important step towards such environmental perception as it allows the car to properly position itself within the road lanes. 
It is also crucial for any subsequent lane departure or trajectory planning decision.
As such, performing accurate camera-based lane detection in real-time is a key enabler of fully autonomous driving. 

Traditional lane detection methods (\eg \cite{Borkar12,Deusch12,Hur13,Jung13,Tan14,Wu14}) rely on a combination of highly-specialized, hand-crafted features and heuristics to identify lane segments.
Popular choices of such hand-crafted cues include color-based features \cite{Chiu05}, the structure tensor \cite{Loose09}, the bar filter \cite{Teng10}, ridge features \cite{Lopez10}, \etc, which are possibly combined with a hough transform \cite{Liu10,Zhou10} and particle or Kalman filters \cite{Kim08,Danescu09,Teng10}.
After identifying the lane segments, post-processing techniques are employed to filter out mis-detections and group segments together to form the final lanes.
For a detailed overview of lane detection systems we refer the reader to \cite{Bar14}.
In general, these traditional approaches are
prone to robustness issues due to road scene variations that can not be easily modeled by such model-based systems.

\begin{figure}[t]
	\begin{center}
		\includegraphics[width=1\linewidth]{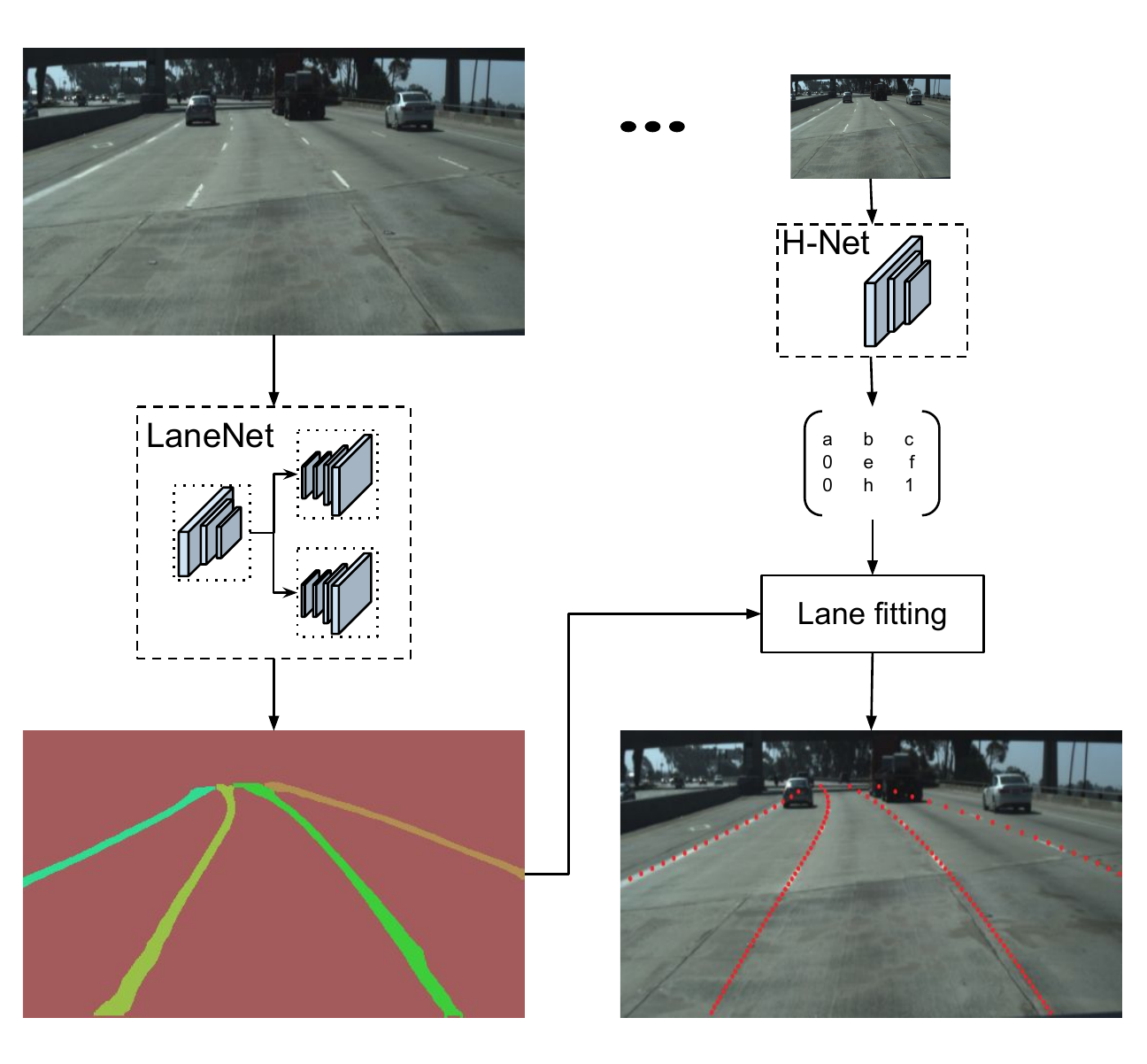}
	\end{center}
	\caption{System overview. Given an input image, LaneNet outputs a lane instance map, by labeling each lane pixel with a lane id. Next, the lane pixels are transformed using the transformation matrix, outputted by H-Net which learns a perspective transformation conditioned on the input image. For each lane a 3rd order polynomial is fitted and the lanes are reprojected onto the image. }
	\label{fig:laneNet_teaser}
\end{figure}

\begin{figure*}[t]
	\begin{center}
		\includegraphics[width=1\linewidth]{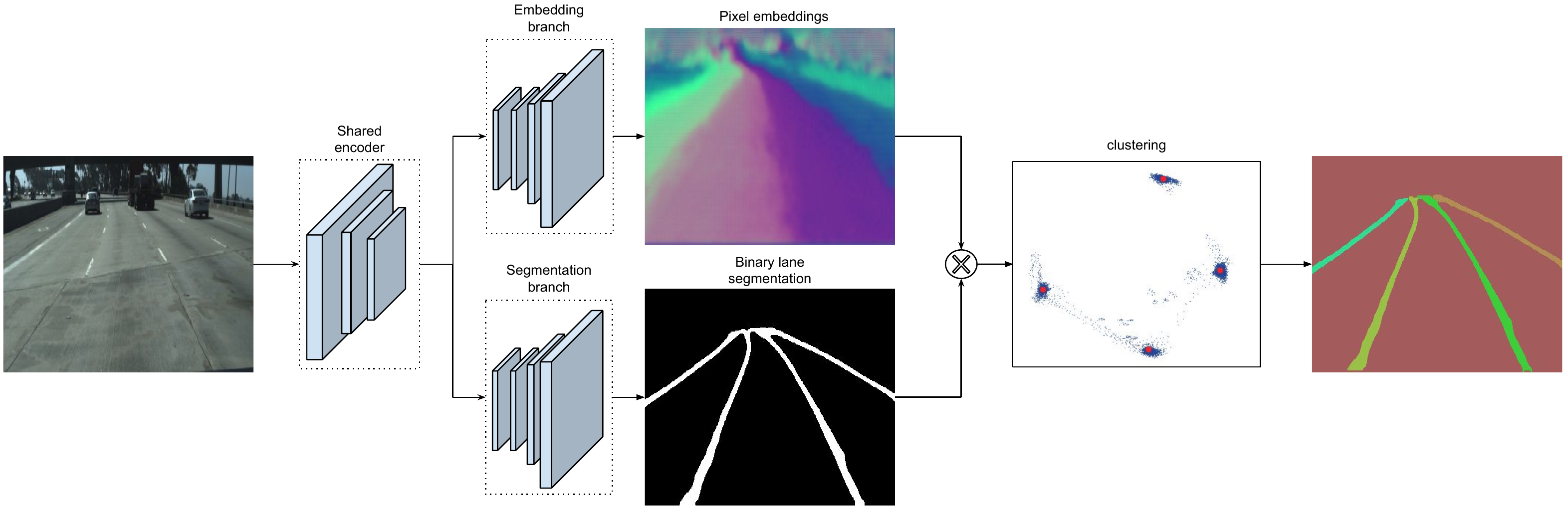}
	\end{center}
	\caption{LaneNet architecture. It consists of two branches. The segmentation branch (bottom) is trained to produce a binary lane mask. The embedding branch (top) generates an N-dimensional embedding per lane pixel, so that embeddings from the same lane are close together and those from different lanes are far in the manifold. For simplicity we show a 2-dimensional embedding per pixel, which is visualized both as a color map (all pixels) and as points (only lane pixels) in a xy grid. After masking out the background pixels using the binary segmentation map from the segmentation branch, the lane embeddings (blue dots) are clustered together and assigned to their cluster centers (red dots).}
	\label{fig:laneNet}
\end{figure*}

More recent methods have replaced the hand-crafted feature detectors with deep networks
to learn dense predictions, \ie pixel-wise lane segmentations. 
Gopalan \etal \cite{Gopalan12} use a pixel-hierarchy feature descriptor to model contextual information and a boosting algorithm to select relevant contextual features for detecting lane markings.
In a similar vein, Kim and Lee \cite{Kim14} combine a Convolutional Neural Network (CNN) with the RANSAC algorithm to detect lanes starting from edge images.
Note that in their method the CNN is mainly used for image enhancement and only if the road scene is complex, \eg it includes roadside trees, fences, or intersections.
Huval \etal \cite{Huval15} show how existing CNN models can be used for highway driving applications, among which an end-to-end CNN that performs lane detection and classification.
He \etal \cite{He16} introduce the Dual-View CNN (DVCNN) that uses a front-view and a top-view image simultaneously to exclude false detections and remove non-club-shaped structures respectively.
Li \etal \cite{Li17} propose the use of a multi-task deep convolutional network that focuses on finding geometric lane attributes, such as location and orientation, together with a Recurrent Neural Network (RNN) that detects the lanes.
Most recently, Lee \etal \cite{Lee17} show how a multi-task network can jointly handle lane and road marking detection and recognition under adverse weather and low illumination conditions.
Apart from the ability of the aforementioned networks to segment out lane markings better \cite{Huval15}, their big receptive field allows them to also estimate lanes even in cases when no markings are present in the image. 
At a final stage, however, the generated binary lane segmentations still need to be disentangled into the different lane instances. 

To tackle this problem, some approaches have applied post-processing techniques that rely again on heuristics, usually guided by geometric properties, as done in \cite{Kim14,Gurghian16} for example.
As explained above, these heuristic methods are computationally expensive and prone to robustness issues due to road scene variations. 
Another line of work \cite{Kim17} casts the lane detection problem as a multi-class segmentation problem, in which each lane forms its own class. 
By doing so, the output of the network contains disentangled binary maps for each lane and can be trained in an end-to-end manner. 
Despite its advantages, this method is limited to detecting only a predefined, fixed number of lanes, \ie the ego-lanes. 
Moreover, since each lane has a designated class, it can not cope with lane changes.

In this paper, we go beyond the aforementioned limitations and propose to cast the lane detection problem as an instance segmentation problem, in which each lane forms its own instance within the lane class.
Inspired by the success of dense prediction networks in semantic segmentation \cite{Long15, Noh15, Ronneberger15, Chen16}
and instance segmentation tasks \cite{Zhang15, Dai16, Romera16, Bai16, He17, Brabandere17},
we design a branched, multi-task network, like \cite{Neven17} for lane instance segmentation, consisting of a lane segmentation branch and a lane embedding branch that can be trained end-to-end.
The lane segmentation branch has two output classes, background or lane, while the lane embedding branch further disentangles the segmented lane pixels into different lane instances.
By splitting the lane detection problem into the aforementioned two tasks, we can fully utilize the power of the lane segmentation branch without it having to assign different classes to different lanes. 
Instead, the lane embedding branch, which is trained using a clustering loss function, assigns a lane id to each pixel from the lane segmentation branch while ignoring the background pixels.
By doing so, we alleviate the problem of lane changes and we can handle a variable number of lanes, unlike \cite{Kim17}.

Having estimated the lane instances, \ie which pixels belong to which lane, as a final step we would like to convert each one of them into a parametric description. 
To this end, curve fitting algorithms have been widely used in the literature.
Popular models are cubic polynomials \cite{Smuda06,Loose09}, splines \cite{Aly08} or clothoids \cite{Gackstatter10}.
To increase the quality of the fit while retaining computational efficiency, it is common to convert the image into a "bird's-eye view" using a perspective transformation~\cite{Bertozzi96} and perform the curve fitting there. 
Note that the fitted line in the bird's-eye view can be reprojected into the original image via the inverse transformation matrix.
Typically, the transformation matrix is calculated on a single image, and kept fixed. 
However, if the ground-plane changes form (\eg by sloping uphill), this fixed transformation is no longer valid. 
As a result, lane points close to the horizon may be projected into infinity, affecting the line fitting in a negative way.

To remedy this situation we also apply a perspective transformation onto the image before fitting a curve, but in contrast to existing methods that rely on a fixed transformation matrix for doing the perspective transformation, we train a neural network to output the transformation coefficients.
In particular, the neural network takes as input the image and is optimized with a loss function that is tailored to the lane fitting problem.
An inherent advantage of the proposed method is that the lane fitting is robust against road plane changes and is specifically optimized for better fitting the lanes. 
An overview of our full pipeline can be seen in Fig.~\ref{fig:laneNet_teaser}.

Our contributions can be summarized to the following:
(1) A branched, multi-task architecture to cast the lane detection problem as an instance segmentation task, that handles lane changes and allows the inference of an arbitrary number of lanes. 
%-(2)
In particular, the
lane segmentation branch outputs dense, per-pixel lane segments, while the lane embedding branch further disentangles the segmented lane pixels into different lane instances.
%-(3)
(2) A network that given the input image estimates the parameters of a perspective transformation that allows for lane fitting robust against road plane changes, \eg up/downhill slope.

The remainder of the paper is organized as follows.
Section~\ref{sec:approach} describes our pipeline for semantic and instance lane segmentation, followed by our approach for converting the segmented lane instances into parametric lines.
Experimental results of the proposed pipeline are presented in Section~\ref{sec:results}. 
Finally, Section~\ref{sec:conclusion} concludes our work.

\section{METHOD}
\label{sec:approach}

We train a neural network end-to-end for lane detection, in a way that copes with the aforementioned problem of lane switching as well as the limitations on the number of lanes. This is achieved by treating lane detection as an instance segmentation problem. The network, which we will refer to as LaneNet (cf. Fig.~\ref{fig:laneNet}), combines the benefits of binary lane segmentation with a clustering loss function designed for one-shot instance segmentation. In the output of LaneNet, each lane pixel is assigned the id of their corresponding lane. This is further explained in the Section~\ref{subsec:lanenet}.

Since LaneNet outputs a collection of pixels per lane, we still have to fit a curve through these pixels to get the lane parametrization. Typically, the lane pixels are first projected into a "bird's-eye view" representation, using a fixed transformation matrix. However, due to the fact that the transformation parameters are fixed for all images, this raises issues when non-flat ground-planes are encountered, \eg in slopes. To alleviate this problem, we train a network, referred to as H-Net, that estimates the parameters of an "ideal" perspective transformation, conditioned on the input image. This transformation is not necessarily the typical "bird's eye view". Instead, it is the transformation in which the lane can be optimally fitted with a low-order polynomial. Section~\ref{subsec:lane_fitting} describes this procedure.

\subsection{LANENET}
\label{subsec:lanenet}

LaneNet is trained end-to-end for lane detection, by treating lane detection as an instance segmentation problem. This way, the network is not constrained on the number of lanes it can detect and is able to cope with lane changes. The instance segmentation task consists of two parts, a segmentation and a clustering part, which are explained in more detail in the following sections. To increase performance, both in terms of speed and accuracy \cite{Neven17}, these two parts are jointly trained in a multi-task network (see Fig.~\ref{fig:laneNet}).

\textbf{binary segmentation}
The segmentation branch of LaneNet (see Fig.~\ref{fig:laneNet}, bottom branch) is trained to output a binary segmentation map, indicating which pixels belong to a lane and which not. To construct the ground-truth segmentation map, we connect all ground-truth lane points\footnote{Depending on the dataset, this can be a discretized set of lane points, parts of lane markings, etc.} together, forming a connected line per lane. Note that we draw these ground-truth lanes even through objects like occluding cars, or also in the absence of explicit visual lane segments, like dashed or faded lanes. This way, the network will learn to predict lane location even when they are occluded or in adverse circumstances. The segmentation network is trained with the standard cross-entropy loss function. Since the two classes (lane/background) are highly unbalanced, we apply bounded inverse class weighting, as described in \cite{Paszke16}.

\textbf{instance segmentation}
To disentangle the lane pixels identified by the segmentation branch, we train the second branch of LaneNet for lane instance embedding (see fig.~\ref{fig:laneNet}, top branch). Most popular detect-and-segment approaches (\eg \cite{He17,Dai16}) are not ideal for lane instance segmentation, since bounding box detection is more suited for compact objects, which lanes are not. Therefore we use a one-shot method based on distance metric learning, proposed by De Brabandere~\etal~\cite{Brabandere17}, which can easily be integrated with standard feed-forward networks and which is specifically designed for real-time applications. 

By using their clustering loss function, the instance embedding branch is trained to output an embedding for each lane pixel so that the distance between pixel embeddings belonging to the same lane is small, whereas the distance between pixel embeddings belonging to different lanes is maximized. By doing so, the pixel embeddings of the same lane will cluster together, forming unique clusters per lane. This is achieved through the introduction of two terms, a variance term ($L_{var}$), that applies a pull force on each embedding towards the mean embedding of a lane, and a distance term ($L_{dist}$), that pushes the cluster centers away from each other. Both terms are hinged: the pull force is only active when an embedding is further than $\delta_v$ from its cluster center, and the push force between centers is only active when they are closer than $\delta_d$ to each-other. With $C$ denoting the number of clusters (lanes), $N_c$ the number of elements in cluster $c$, $x_i$ a pixel embedding, $\mu_c$ the mean embedding of cluster $c$, $\lVert \cdot \rVert$ the L2 distance, and $\left[ x \right]_{+} = \textrm{max}(0,x)$ the hinge, the total loss $L$ is equal to $L_{var} + L_{dist}$ with: 

\begin{equation}
\begin{cases}
L_{var} = \frac{1}{C} \sum_{c=1}^{C} \frac{1}{N_c} \sum_{i=1}^{N_c} \left[ \lVert \mu_c - x_i \rVert - \delta_{\textrm{v}} \right]_{+}^2 \\ \\
L_{dist} = \frac{1}{C (C-1)} \mathop{\sum_{c_A = 1}^{C} \sum_{c_B = 1,}^{C}}_{c_A \neq c_B} \left[\delta_{\textrm{d}} - \lVert \mu_{c_A} - \mu_{c_B} \rVert \right]_{+}^2 \\ 
%L_{reg} = \frac{1}{C} \sum_{c=1}^{C} \lVert \mu_{c} \rVert 
\end{cases}
\end{equation}

Once the network has converged, the embeddings of lane pixels will be clustered together (see fig.~\ref{fig:laneNet}), so that each cluster will lay further than $\delta_d$ from each other and the radius of each cluster is smaller than $\delta_v$.

\textbf{clustering} The clustering is done by an iterative procedure. By setting $\delta_d > 6 \delta_v$ in the above loss, one can take a random lane embedding and threshold around it with a radius of $2\delta_v$ to select all embeddings belonging to the same lane. This is repeated until all lane embeddings are assigned to a lane. To avoid selecting an outlier to threshold around, we first use mean shift to shift closer to the cluster center and then do the thresholding (see Fig.~\ref{fig:laneNet}). 

\textbf{network architecture} LaneNet's architecture is based on the encoder-decoder network ENet~\cite{Paszke16}, which is consequently modified into a two-branched network. Since ENet's encoder contains more parameters than the decoder, completely sharing the full encoder between the two tasks would lead to unsatisfying results \cite{Neven17}. As such, while the original ENet's encoder consists of three stages (stage 1,2,3), LaneNet only shares the first two stages (1 and 2) between the two branches, leaving stage 3 of the ENet encoder and the full ENet decoder as the backbone of each separate branch. The last layer of the segmentation branch outputs a one channel image (binary segmentation), whereas the last layer of the embedding branch outputs a N-channel image, with N the embedding dimension. This is schematically depicted in Fig.~\ref{fig:laneNet}. Each branch's loss term is equally weighted and back-propagated through the network.

\begin{figure}[t]
	\begin{center}
		\includegraphics[width=1\linewidth]{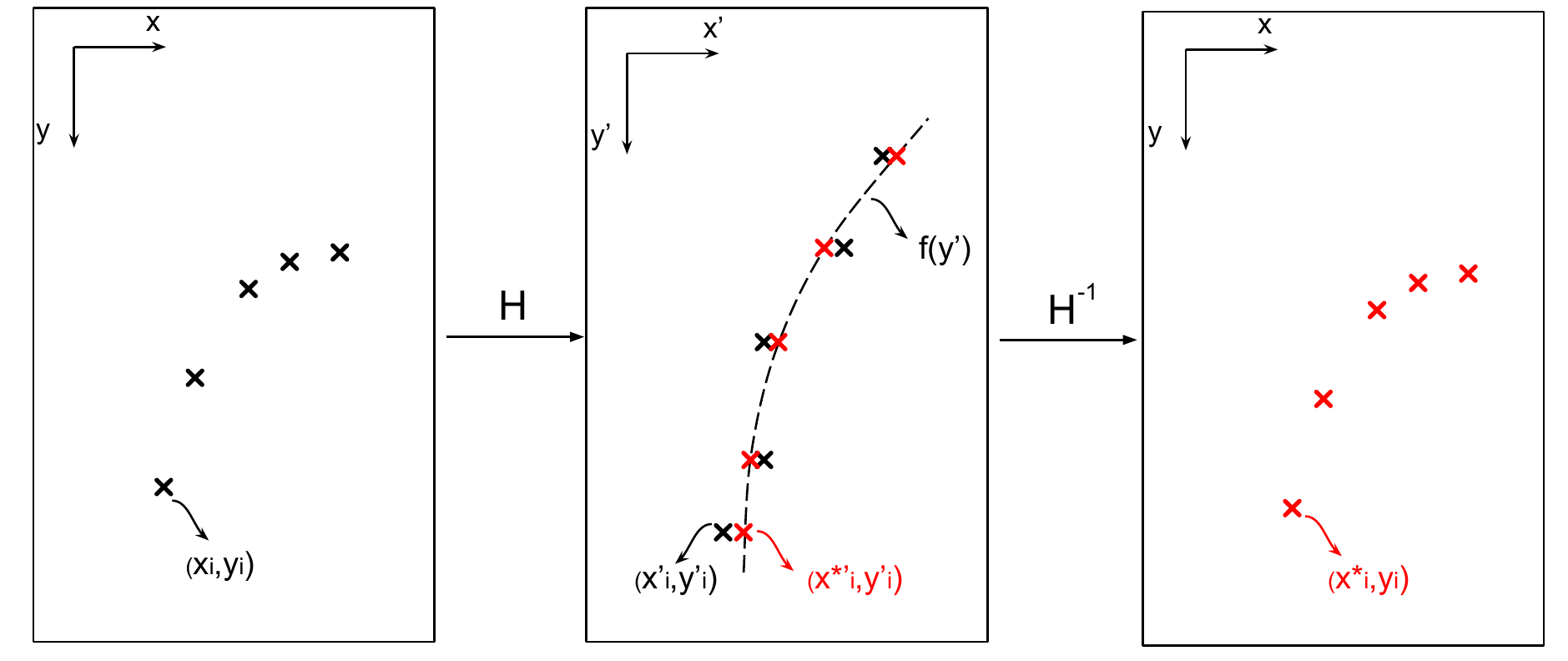}
	\end{center}
	\caption{Curve fitting. 
    {\it Left:}
    The lane points are transformed using the matrix H generated by H-Net. 
    {\it Mid:}
    A line is fitted through the transformed points and the curve is evaluated at different heights (red points). 
    {\it Right:}
    The evaluated points are transformed back to the original image space.}
	\label{fig:transformation_explanation}
\end{figure}

\subsection{CURVE FITTING USING H-NET}
\label{subsec:lane_fitting}

As explained in the previous section, the output of LaneNet is a collection of pixels per lane. Fitting a polynomial through these pixels in the original image space is not ideal, as one has to resort to higher order polynomials to be able to cope with curved lanes. A frequently used solution to this problem is to project the image into a "bird's-eye view" representation, in which lanes are parallel to each other and as such, curved lanes can be fitted with a 2nd to 3rd order polynomial. 

\begin{figure}
	\begin{center}
		\includegraphics[width=1\linewidth]{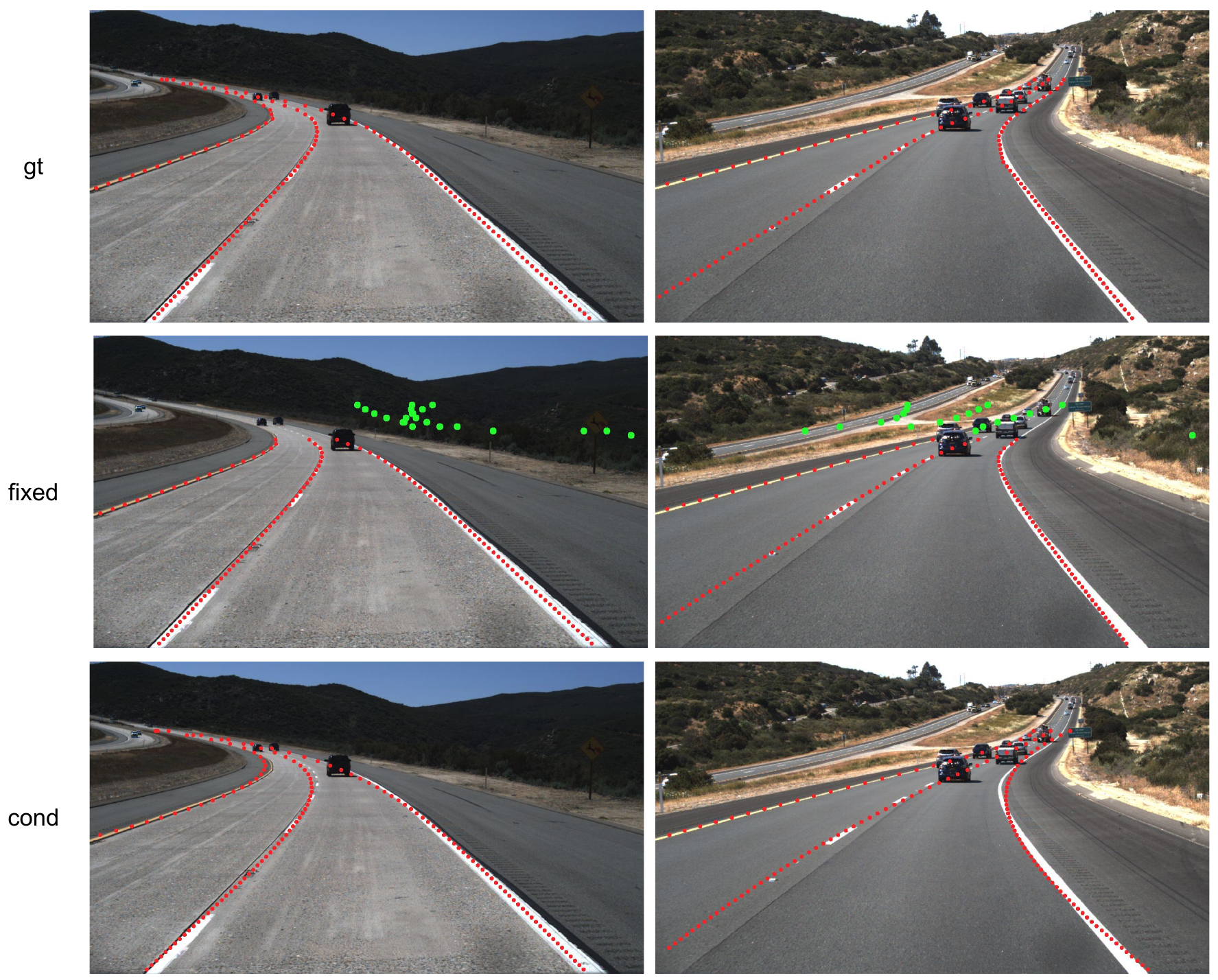}
	\end{center}
	\caption{Comparison between a fixed homography and a conditional homography (using H-Net) for lane fitting. The green dots can't be fitted correctly using a fixed homography because of groundplane changes, which can be resolved by using a conditional homography using H-Net (last row).}
	\label{fig:comp_transf}
\end{figure}

However, in these cases the transformation matrix H is calculated once, and kept fixed for all images. Typically, this leads to errors under ground-plane changes where the vanishing-point, which is projected onto infinity, shifts up or downwards (see Fig. \ref{fig:comp_transf}, second row).

To resolve this issue we train a neural network, H-Net, with a custom loss function: the network is optimized end-to-end to predict the parameters of a perspective transformation H, in which the transformed lane points can be optimally fitted with a 2nd or 3rd order polynomial. The prediction is conditioned on the input image, allowing the network to adapt the projection parameters under ground-plane changes, so that the lane fitting will still be correct (see the last row of Fig.~\ref{fig:comp_transf}). In our case, H has 6 degrees of freedom: 

$$
\mathrm{H}=
\begin{bmatrix}
a & b & c \\
0 & d & e \\
0 & f & 1 \\
\end{bmatrix}
$$

The zeros are placed to enforce the constraint that horizontal lines remain horizontal under the transformation. 

\textbf{curve fitting} Before fitting a curve through the lane pixels $\mathbf{P}$, the latter are transformed using the transformation matrix outputted by H-Net. Given a lane pixel $\mathbf{p_i} = [x_i,y_i,1]^T \in \mathbf{P}$, the transformed pixel $\mathbf{p'_i} = [x'_i,y'_i,1]^T \in \mathbf{P'}$ is equal to $\mathrm{H}\mathbf{p_i}$. Next, the least-squares algorithm is used to fit a n-degree polynomial, $f(y')$, through the transformed pixels $\mathbf{P'}$. 

To get the x-position, $x^*_i$ of the lane at a given y-position $y_i$, the point $\mathbf{p_i}=[-,y_i,1]^T$ is transformed to $\mathbf{p'_i}=\mathrm{H}\mathbf{p_i}=[-,y'_i,1]^T$ and evaluated as: $x'^*_i=f(y'_i)$. Note that the x-value is of no importance and indicated with '-'. By re-projecting this point $\mathbf{p'^*_i}=[x'^*_i,y'_i,1]^T$ into the original image space we get:  $\mathbf{p^*_i}=\mathrm{H^{-1}}\mathbf{p'^*_i}$ with $\mathbf{p^*_i}=[x^*_i,y_i,1]^T$.
This way, we can evaluate the lane at different y positions. This process is illustrated in fig.~\ref{fig:transformation_explanation}.

\textbf{loss function} In order to train H-Net for outputting the transformation matrix that is optimal for fitting a polynomial through lane pixels, we construct the following loss function. Given $N$ ground-truth lane points $\mathbf{p_i}=[x_i,y_i,1]^T \in \mathbf{P}$, we first transform these points using the output of H-Net:
$$\mathbf{P'}= \mathrm{H}\mathbf{P}$$
with $\mathbf{p'_i} = [x'_i,y'_i,1]^T \in \mathbf{P'}$.
Through these projected points, we fit a polynomial $f(y') = \alpha y'^2 + \beta y' + \gamma$ using the least squares closed-form solution:
$$\mathbf{w} = (\mathbf{Y}^T\mathbf{Y})^{-1}\mathbf{Y}^T\mathbf{x'}$$
with $\mathbf{w}=[\alpha, \beta, \gamma]^T$ , $\mathbf{x}'=[x'_1, x'_2, ... , x'_N]^T$ and 
$$	\mathbf{Y} = 
	\begin{bmatrix}
    	y_1^{'2} & y'_1 & 1 \\
        \vdots & \vdots & \vdots \\
        y_N^{'2} & y'_N & 1 \\
    \end{bmatrix}
$$
for the case of a 2nd order polynomial. The fitted polynomial is evaluated at each $y'_i$ location, giving us a $x'^*_i$ prediction. 
These predictions are projected back: $\mathbf{p^*_i}=\mathrm{H^{-1}}\mathbf{p'^*_i}$ with $\mathbf{p^*_i}=[x^*_i,y_i,1]^T$ and $\mathbf{p'^*_i}=[x'^*_i,y'_i,1]^T$.
The loss is: 

$$Loss = \frac{1}{N}\sum_{i=1,N}(x^*_i - x_i)^2$$

Since the lane fitting is done by using the closed-form solution of the least squares algorithm, the loss is differentiable. We use automatic differentiation to calculate the gradients. 

\textbf{network architecture} The network architecture of H-Net is kept intentionally small and is constructed out of consecutive blocks of 3x3 convolutions, batchnorm and ReLUs. The dimension is decreased using max pooling layers, and in the end 2 fully-connected layers are added. See Table~\ref{tab:hnet_network} for the complete network structure.

\begin{table}
\begin{center}
\begin{tabular}{c|c|c|c}
  	Type & Filters & Size/Stride & Output \\
  	\hline
	Conv+BN+ReLU & 16 & 3x3 & 128x64 \\
    Conv+BN+ReLU & 16 & 3x3 & 128x64 \\
 	Maxpool & & 2x2/2 & 64x32 \\
 	Conv+BN+ReLU & 32 & 3x3 &  64x32 \\   
	Conv+BN+ReLU & 32 & 3x3 & 64x32 \\
    Maxpool & & 2x2/2 & 32x16 \\
	Conv+BN+ReLU & 64 & 3x3 & 32x16 \\
	Conv+BN+ReLU & 64 & 3x3 & 32x16 \\
    Maxpool & & 2x2/2 & 16x8 \\   
    Linear+BN+ReLU & & 1x1 & 1024 \\
    \hline
    Linear & & 1x1 & 6 \\
\end{tabular}
\end{center}
\caption{H-Net network architecture.}
\label{tab:hnet_network}
\end{table}

\section{RESULTS}
\label{sec:results}

\begin{figure*}[t]
	\begin{center}
		\includegraphics[width=1\linewidth]{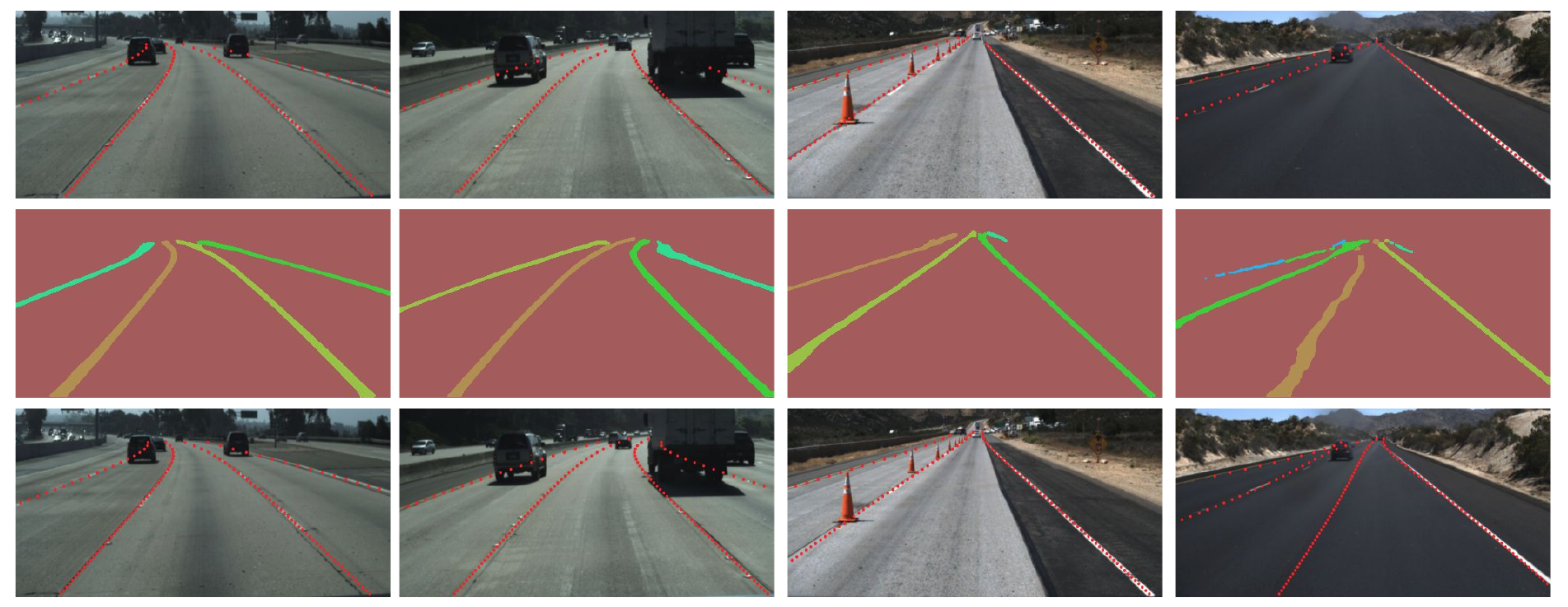}
	\end{center}
	\caption{Visual results. \emph{Top row}: ground-truth lane points. \emph{Middle row}: LaneNet output. \emph{Bottom row}: final lane predicts after lane fitting.}
	\label{fig:vis_results}
\end{figure*}

\subsection{Dataset}

At the moment, the tuSimple lane dataset~\cite{tusimple} is the only large scale dataset for testing deep learning methods on the lane detection task. It consists of 3626 training and 2782 testing images, under good and medium weather conditions. They are recorded on 2-lane/3-lane/4-lane or more highway roads, at different daytimes. For each image, they also provide the 19 previous frames, which are not annotated. The annotations come in a json format, indicating the x-position of the lanes at a number of discretized y-positions. On each image, the current (ego) lanes and left/right lanes are annotated and this is also expected on the test set. When changing lanes, a 5th lane can be added to avoid confusion. 

The accuracy is calculated as the average correct number of points per image: 
$$acc = \sum_{im} \frac{C_{im}}{S_{im}}$$
with $C_{im}$ the number of correct points and $S_ {im}$ the number of ground-truth points. A point is correct when the difference between a ground-truth and predicted point is less than a certain threshold. Together with the accuracy, they also provide the false positive and false negative scores:
$$ FP=\frac{F_{pred}}{N_{pred}} $$
$$ FN=\frac{M_{pred}}{N_{gt}} $$
with $F_{pred}$ the number of wrongly predicted lanes, $N_{pred}$ the number of predicted lanes, $M_{pred}$ the number of missed ground-truth lanes and $N_{gt}$ the number of all ground-truth lanes. 

\subsection{Setup}

\textbf{LaneNet} is trained with an embedding dimension of 4, with $\delta_v=0.5$ and $\delta_d=3$. The images are rescaled to 512x256 and the network is trained using Adam with a batch size of 8 and a learning rate 5e-4 until convergence. 

\textbf{H-Net} is trained for a 3rd-order polynomial fit, with a scaled version of input image with dimension 128x64. The network is trained using Adam with a batch size of 10 and learning rate 5e-5 until convergence. 

\textbf{Speed} Given an input resolution of 512x256, a 4-dimensional embedding per pixel and using a 3rd order polynomial fit, our lane detection algorithm can run up to 50 frames per second. A full breakdown of the different components can be found in Table~\ref{tab:speed}.

\begin{table}
	\begin{center}
    \begin{tabular}{l|c|c|c}
    	& acc & FP & FN \\
		\hline
        leonardoli & 96.9 & 0.0442 & 0.0197 \\
        XingangPan & 96.5 & 0.0617 & 0.0180 \\
        aslarry & 96.5 & 0.0851 & 0.0269 \\
        dpantoja & 96.2 & 0.2358 & 0.0362 \\
        xxxxcvcxxxx & 96.1 & 0.2033 & 0.0387 \\
        \hline
		ours & 96.4 & 0.0780 & 0.0244 \\
    \end{tabular}
    \end{center}
    \caption{Lane detection performance on the tuSimple test set.}
    \label{tab:results_tusimple}
\end{table}

\begin{table}
\begin{center}
\begin{tabular}{l|c|c|c}
	& 2th ordr (MSE) & 3rd ordr (MSE)& Avg. miss/lane \\
    \hline
    no transform & 53.91 & 17.23 & 0 \\
    fixed transform & 48.09 & 9.42 & 0.105 \\
    cond. transform & 33.82 & \textbf{5.99} & 0 \\
\end{tabular}
\end{center}
\caption{MSE (in pixels) between fitted lane points and gt points on validation set using a 2nd order and 3rd order polynomial under different transformations. A point which can not be fitted is not added to the MSE, but is considered a miss and contributes to the average miss/lane. }
\label{tab:hnet_performance}
\end{table}

\subsection{Experiments}

\begin{table}
\begin{center}
\begin{tabular}{ll|c|c}
           							& & time (ms) 	& fps 	\\ \hline
\multirow{2}{*}{\textbf{LaneNet}}				& Forward pass 		& 12        	& \multirow{2}{*}{62.5}  \\
                       							& Clustering   		& 4.6        	&     	\\ \hline
\multirow{2}{*}{\textbf{H-Net}}         		& Forward pass 		& 0.4        	& \multirow{2}{*}{416.6}   	\\ 
             									& Lane Fitting 		& 2      		&     	\\ \hline
                    							& \multicolumn{1}{r|}{\textbf{Total}}   & \textbf{19}	& \textbf{52.6}  \\     
\end{tabular}    
\end{center}
\caption{Speed of the different components for an image size of 512x256 measured on a Nvidia 1080 Ti. In total, lane detection can run at 52 fps.}
\label{tab:speed}
\end{table}

\textbf{Interpolation method} In Table~\ref{tab:hnet_performance} we calculate the accuracy of lane fitting using no transformation, a fixed transformation and a conditional transformation based on H-Net. We also measure the difference between a 2nd or 3rd order polynomial fit. When directly fitting the curve in the original image space without a transformation, this leads to inferior results; expectedly since curved lanes are difficult to fit using low-order polynomials. 

By using a fixed transformation we already obtain better results, with a 3rd order polynomial performing better than a 2nd order one. However, as already mentioned in Section~\ref{subsec:lane_fitting}, not all lane-points can be fitted under a fixed transformation (see also Fig.~\ref{fig:comp_transf}). When the slope of the ground-plane changes, points close to the vanishing-point can not be fitted correctly and are therefore ignored in the MSE-measure, but still counted as missed points. 

Using the transformation matrix generated by H-Net, which is optimized for lane fitting, the results outperform the lane fitting with a fixed transformation. Not only do we get a better MSE-score, but using this method allows us to fit all points, no matter if the slope of the ground-plane changes.

\textbf{tuSimple results} By using LaneNet combined with a 3rd order polynomial fitting and the transformation matrix from H-Net, we reach 4th place on the tuSimple challenge, with only a 0.5\% difference between the first entry. The results can be seen in Table~\ref{tab:results_tusimple}. Note that we have only trained on the training images of the tuSimple dataset, which is unclear for the other entries, as is their speed performance too.

\section{CONCLUSION}
\label{sec:conclusion}

In this paper we have presented a method for end-to-end lane detection at 50 fps. Inspired by recent instance segmentation techniques, our method can detect a variable number of lanes and can cope with lane change maneuvers, in contrast to other related deep learning approaches. 

In order to parametrize the segmented lanes using low order polynomials, we have trained a network to generate the parameters of a perspective transformation, conditioned on the image, in which lane fitting is optimal. This network is trained using a custom loss function for lane fitting. Unlike the popular "bird's-eye view" approach, our method is robust against ground-plane's slope changes, by adapting the parameters for the transformation accordingly.

\noindent {\bf Acknowledgement:} The work was supported by Toyota, and was carried out at the TRACE Lab at KU Leuven (Toyota Research on Automated Cars in Europe - Leuven).


\begin{thebibliography}{99}

\bibitem{Aly08} M. Aly, Real time Detection of Lane Markers in Urban Streets. Intelligent Vehicles Symposium, pp. 7-12, 2008.
\bibitem{Bai16} M. Bai, R. Urtasun, Deep  watershed  transform  for instance  segmentation. CoRR abs/1611.08303, 2016.
\bibitem{Bar14} A. Bar-Hillel, R. Lerner, D. Levi, G. Raz, Recent progress in road and lane detection: a survey. Mach. Vis. Appl., vol. 25, no. 3, pp. 727-745, 2014.
\bibitem{Borkar12} A. Borkar, M. Hayes, M. T. Smith, A Novel Lane Detection System With Efficient Ground Truth Generation. IEEE Trans. Intelligent Transportation Systems, vol. 13, no. 1, pp. 365-374, 2012.
\bibitem{Brabandere17} B. De Brabandere, D. Neven, L. Van Gool. CoRR abs/1708.02551, 2017.
\bibitem{Chen16} L. Chen, G. Papandreou, I. Kokkinos, K. Murphy, A. Yuille, Deeplab: Semantic image segmentation with deep convolutional nets, atrous convolution, and fully connected crfs. CoRR abs/1606.00915, 2016.
\bibitem{Chiu05} K.-Y. Chiu, S.-F. Lin, Lane detection using color-based segmentation. Intelligent Vehicles Symposium, pp. 706-711, 2005.
\bibitem{Danescu09} R. Danescu, S. Nedevschi, Probabilistic Lane Tracking in Difficult Road Scenarios Using Stereovision. IEEE Trans. Intelligent Transportation Systems, vol. 10, no. 2, pp. 272-282, 2009.
\bibitem{Deusch12} H. Deusch, J. Wiest, S. Reuter, M. Szczot, M. Konrad, K. Dietmayer, A random finite set approach to multiple lane detection. ITSC, pp. 270-275, 2012.
\bibitem{Gackstatter10} C. Gackstatter, P. Heinemann, S. Thomas, G. Klinker, Stable road lane model based on clothoids. Advanced Microsystems for Automotive Applications, pp. 133-143, 2010.
\bibitem{Gopalan12} R. Gopalan, T. Hong, M. Shneier, R. Chellappa, A Learning Approach Towards Detection and Tracking of Lane Markings. IEEE Trans. Intelligent Transportation Systems, vol. 13, no. 3, pp. 1088-1098, 2012.
\bibitem{Gurghian16} A. Gurghian, T. Koduri, S. V. Bailur, K. J. Carey, V. N. Murali, DeepLanes: End-To-End Lane Position Estimation Using Deep Neural Networks. CVPR Workshops, pp. 38-45, 2016.
\bibitem{He16} B. He, R. Ai, Y. Yan, X. Lang, Accurate and robust lane detection based on Dual-View Convolutional Neutral Network. Intelligent Vehicles Symposium, pp. 1041-1046, 2016.
\bibitem{He17} K. He, G. Gkioxari, P. Dollar, R. Girshick, Mask R-CNN. CoRR abs/1703.06870, 2017.
\bibitem{Hur13} J. Hur, S.-N. Kang, S.-W. Seo, Multi-lane detection in urban driving environments using conditional random fields. Intelligent Vehicles Symposium, pp. 1297-1302, 2013.
\bibitem{Huval15} B. Huval, T. Wang, S. Tandon, J. Kiske, W. Song, J. Pazhayampallil, M. Andriluka, P. Rajpurkar, T. Migimatsu, R. Cheng-Yue, F. Mujica, A. Coates, A. Y. Ng, An Empirical Evaluation of Deep Learning on Highway Driving. CoRR abs/1504.01716, 2015.
\bibitem{Jung13} H. Jung, J. Min, J. Kim, An efficient lane detection algorithm for lane departure detection. Intelligent Vehicles Symposium, pp. 976-981, 2013.
\bibitem{Kim08} Z. Kim, Robust Lane Detection and Tracking in Challenging Scenarios. IEEE Trans. Intelligent Transportation Systems, vol. 9, no. 1, pp. 16-26, 2008.
\bibitem{Kim14} J. Kim, M. Lee, Robust Lane Detection Based On Convolutional Neural Network and Random Sample Consensus. ICONIP, pp. 454-461, 2014.
\bibitem{Kim17} J. Kim, C. Park, End-To-End Ego Lane Estimation Based on Sequential Transfer Learning for Self-Driving Cars. CVPR Workshops, pp. 1194-1202, 2017.
\bibitem{Lee17} S. Lee, J.-S. Kim, J. S. Yoon, S. Shin, O. Bailo, N. Kim, T.-H. Lee, H. S. Hong, S.-H. Han, I. S. Kweon, VPGNet: Vanishing Point Guided Network for Lane and Road Marking Detection and Recognition. CoRR abs/1710.06288, 2017.
\bibitem{Li17} J. Li, X. Mei, D. V. Prokhorov, D. Tao, Deep Neural Network for Structural Prediction and Lane Detection in Traffic Scene. IEEE Trans. Neural Netw. Learning Syst., vol. 28, no.3, pp. 690-703, 2017.
\bibitem{Liu10} G. Liu, F. W{\"o}rg{\"o}tter, I. Markelic, Combining Statistical Hough Transform and Particle Filter for robust lane detection and tracking. Intelligent Vehicles Symposium, pp. 993-997, 2010.
\bibitem{Long15} J. Long, E. Shelhamer, T. Darrell, Fully convolutional networks for semantic segmentation. CVPR 2015.
\bibitem{Loose09} H. Loose, U. Franke, C. Stiller, Kalman particle filter for lane recognition on rural roads. Intelligent Vehicles Symposium, pp. 60-65, 2009.
\bibitem{Lopez10} A. L{\'o}pez, J. Serrat, C. Canero, F. Lumbreras, T. Graf, Robust lane markings detection and road geometry computation. International Journal of Automotive Technology, vol. 11, no. 3, pp. 395-407, 2010.
\bibitem{Neven17} D. Neven, B. De Brabandere, S. Georgoulis, M. Proesmans, L. Van Gool, Fast Scene Understanding for Autonomous Driving. Deep Learning for Vehicle Perception, workshop at the IEEE Symposium on Intelligent Vehicles, 2017.
\bibitem{Noh15} H. Noh, S. Hong, B. Han, Learning deconvolution network for semantic segmentation. ICCV 2015.
\bibitem{Paszke16} A. Paszke, A. Chaurasia, S. Kim, E. Culurciello, ENet: A deep neural network architecture for real-time semantic segmentation. CoRR abs/1606.02147, 2016.
\bibitem{Romera16} B. Romera-Paredes, P. H. Torr, Recurrent instance segmentation. ECCV, 2016.
\bibitem{Ronneberger15} O. Ronneberger, P. Fischer, T. Brox, U-net: Convolutional networks for biomedical image segmentation. International Conference on Medical Image Computing and Computer-Assisted Intervention, pp. 234-241, 2015.
\bibitem{Smuda06} P. Smuda, R. Schweiger, H. Neumann, W. Ritter, Multiple cue data fusion with particle filters for road course detection in vision systems. Intelligent Vehicles Symposium, pp. 400-405, 2006.
\bibitem{Tan14} H. Tan, Y. Zhou, Y. Zhu, D. Yao, K. Li, A novel curve lane detection based on Improved River Flow and RANSA. ITSC, pp. 133-138, 2014.
\bibitem{Teng10} Z. Teng, J.-H. Kim, D.-J. Kang, Real-time Lane detection by using multiple cues. Control Automation and Systems, pp. 2334-2337, 2010.
\bibitem{Wu14} P.-C. Wu, C.-Y. Chang, C.-H. Lin, Lane-mark extraction for automobiles under complex conditions. Pattern Recognition, vol. 47, no. 8, pp. 2756-2767, 2014.
\bibitem{Zhang15} Z. Zhang, A. G. Schwing, S. Fidler, R. Urtasun, Monocular  object  instance  segmentation  and  depth  ordering  with
CNNs. ICCV, 2015.
\bibitem{Zhou10} S. Zhou, Y. Jiang, J. Xi, J. Gong, G. Xiong, H. Chen, A novel lane detection based on geometrical model and Gabor filter. Intelligent Vehicles Symposium, pp. 59-64, 2010.
\bibitem{Dai16} J. Dai, K. He, and J. Sun.  Instance-aware semantic segmentation via multi-task network cascades. In CVPR, 2016
\bibitem{Bertozzi96} M. Bertozzi and A. Broggi. Real-time lane and obstacle detection on the system.IV, 1996
\bibitem{tusimple} The tuSimple lane challange, http://benchmark.tusimple.ai/

\end{thebibliography}
\end{document}